%% file: paper1242.tex
\documentclass[runningheads]{llncs}
\usepackage{graphicx}

\begin{document}
\title{Domain Adaptive Nuclei Instance Segmentation and Classification via Category-aware Feature Alignment and Pseudo-labelling}
\titlerunning{CAPL-Net}
\author{Canran Li\inst{1} \and
Dongnan Liu\inst{1} \and
Haoran Li\inst{2,3} \and
Zheng Zhang\inst{4} \and
Guangming Lu\inst{4} \and
Xiaojun Chang\inst{2} \and
Weidong Cai\inst{1}}
\authorrunning{C. Li et al.}
\institute{School of Computer Science, University of Sydney, Australia\\ \and
ReLER, AAII, University of Technology Sydney, Australia \\ \and
Department of Data Science and Artificial Intelligence, Monash University, Australia\\ \and
School of Computer Science and Technology, Harbin Institute of Technology, Shenzhen, China \\
\email{cali5184@uni.sydney.edu.au}\\
%\email{\{dongnan.liu,tom.cai\}@sydney.edu.au} \\
%\email{haoran.li@monash.edu}\\
%\email{darrenzz219@gmail.com} \\
%\email{luguangm@hit.edu.cn} \\
%\email{Xiaojun.Chang@uts.edu.au}
}
\maketitle              % typeset the header of the contribution
\begin{abstract}
Unsupervised domain adaptation (UDA) methods have been broadly utilized to improve the models' adaptation ability in general computer vision. However, different from the natural images, there exist huge semantic gaps for the nuclei from different categories in histopathology images. It is still under-explored how could we build generalized UDA models for precise segmentation or classification of nuclei instances across different datasets. In this work, we propose a novel deep neural network, namely Category-Aware feature alignment and Pseudo-Labelling Network (CAPL-Net) for UDA nuclei instance segmentation and classification. Specifically, we first propose a category-level feature alignment module with dynamic learnable trade-off weights. Second, we propose to facilitate the model performance on the target data via self-supervised training with pseudo labels based on nuclei-level prototype features. Comprehensive experiments on cross-domain nuclei instance segmentation and classification tasks demonstrate that our approach outperforms state-of-the-art UDA methods with a remarkable margin.

\end{abstract}
\section{Introduction}

Automatic nuclei instance segmentation and classification are crucial for digital pathology with various application scenarios, such as tumour classification and cancer grading \cite{1}. However, manual labelling is limited by high subjective, low reproducibility, and resource-intensive~\cite{2,3}. Although deep learning-based methods can achieve appealing nuclei recognition performance, they require sufficient labelled data for training \cite{3,4,5,6,7}. By directly adopting the off-the-shelf deep learning models to a new histopathology dataset with a distinct distribution, the models suffer from performance drop due to domain bias \cite{8,9}. Recently, unsupervised domain adaption methods have been proposed to tackle this issue \cite{10,11,12}, and enable the learning models to transfer the knowledge from one labelled source domain to the other unlabelled target domain \cite{13,14,15}.

Several methods have recently been proposed for unsupervised domain adaptive nuclei instance segmentation in histopathology images \cite{8,9}. Inspired by CyCADA \cite{16}, Liu et al. \cite{8} first synthesize target-like images and use a nuclei inpainting mechanism to remove the incorrectly synthesized nuclei. The adversarial training strategies are then used separately at the image-, semantic- and instance-level with a task re-weighting mechanism. However, this work can only address the instance segmentation for nuclei within the same class. In Yang et al.'s work \cite{9} , firstly, local features are aligned by an adversarial domain discriminator, and then a pseudo-labelling self-training approach is used to further induce the adaptation. However, the performance improvement of this work relies on weak labels and fails to get good training results without target domain labels. In addition, although this work is validated on the cross-domain nuclei segmentation and classification, their adaptation strategies are class-agnostic. In the real clinical, the nuclei objects in the histopathology images belong to various classes, and the characteristics of the nuclei in different classes are also distinct~\cite{17}. In addition, the number of objects within each category is also imbalanced in the histopathology datasets~\cite{18}. To this end, previous UDA methods are limited for cross-domain nuclei segmentation and classification due to the lack of analysis on the nuclei classes when transferring the knowledge.

To address the aforementioned issues, in this work, we study cross-domain nuclei instance segmentation and classification via a novel class-level adaptation framework. First, we propose a category-aware feature alignment module to facilitate the knowledge transfer for the cross-domain intra-class features while avoiding negative transfer for the inter-class ones. Second, a self-supervised learning stage via nuclei-level feature prototypes is further designed to improve the model performance on the unlabelled target data. Extensive experiments indicated the effectiveness of our proposed method by outperforming state-of-the-art UDA methods on nuclei instance segmentation and classification tasks. Furthermore, the performance of our UDA method is comparable or even better than the fully-supervised upper bound under various metrics.

\section{Methods}
\subsection{Overview}

Our proposed model is based on Hover-Net \cite{6}, a state-of-the-art method for fully-supervised nuclei instance segmentation and classification. The framework is constructed by three branches sharing the same encoder for feature extraction and using three decoders for different tasks: 1) Nuclear pixel (NP) branch to perform binary classification of a pixel (nuclei or background); 2) Hover (HV) branch to predict the horizontal and vertical distances of nuclei pixels to their centroid; 3) Nuclear classiﬁcation (NC) branch to classify the nuclei types of pixels. The supervised Hover-Net loss function of our model is deﬁned as $\mathcal{L}_F$:

\begin{equation}
\mathcal{L}_F =  \mathcal{L}_{np} + \mathcal{L}_{hover} +\mathcal{L}_{nc} 
\end{equation}

\begin{figure}
\center{\includegraphics[scale=0.28]{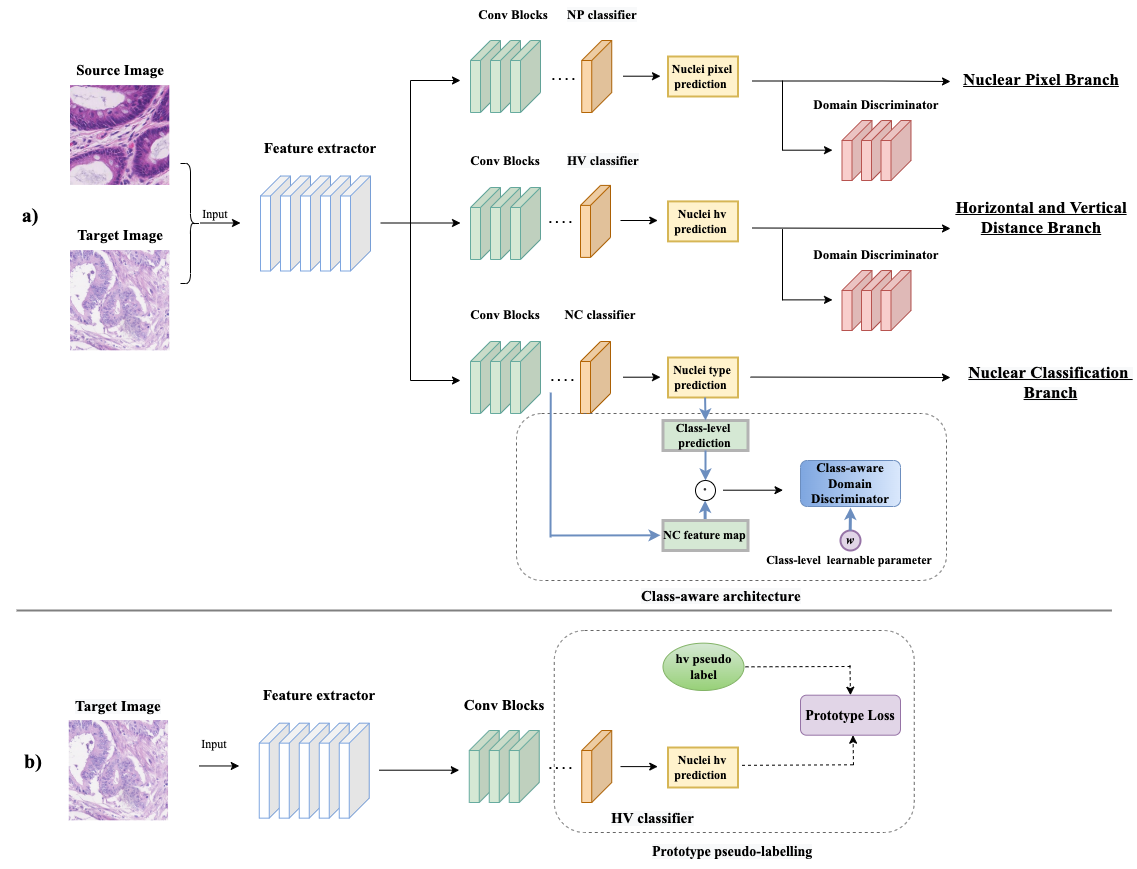}}
\caption{Overview of the proposed category-aware prototype pseudo-labelling network.} \label{fig1} \label{framework}
\end{figure}

The network architecture of our proposed model is shown in
Fig.~\ref{fig1}. Our proposed UDA framework is optimized in two stages. First, class-level feature alignment modules are proposed to alleviate the domain gap at the feature level.
In the second stage, the pseudo-labelling process enhanced by the nuclei-level prototype is further proposed for self-supervised learning on the unlabelled target images.

\subsection{Category-aware Feature Alignment}

In the histopathology datasets with multi-class nuclei, there is a very large gap in the number of nuclear in each category. This may lead to many images in the dataset having only a few nuclear categories, and some nuclear types are absent from an image. In addition, the features of the nuclei from different classes also vary. Under this situation, the typical class-agnostic feature alignment strategies may lead to the negative transfer of the features in different categories. To tackle this issue, for the NC branch on nuclei classification, we propose to conduct feature alignment for the cross-domain features within each class separately. Particularly, an adversarial domain discriminator $D_{c}$ is introduced to adapt the features under the class $c$. Compared with directly employing a class-agnostic domain discriminator for the NC branch, the class-aware domain discriminators can avoid the misalignment across different classes and further encourage the knowledge transfer in classification learning. 

The detailed paradigm is shown in Fig.~\ref{fig1}a. First, we denote the features of the NC branch in the typical Hover-Net as $F_{nc}$, and the N-class prediction as $P_{nc}$. Note that $P_{nc}$ contains $N$ channels, and the predictions in each channel represent the classification results for each specific class. For each category $c$, we formulate the $P_{nc}$ into a binary class prediction map $P_{nc}^{c}$, where the pixel value is set to 1 if it belongs to the nuclei in this category, otherwise set to 0. To incorporate the class-aware information into the feature alignment, we propose to generate the prototype features $F_{pt}^{c}$ for the class $c$ by dot-multiplying $F_{nc}$ with the binary $P_{nc}^{c}$. In addition, if any $P_{nc}^{c}$ is empty, we will not perform subsequent training on the nuclei in this class. In other words, we only deal with the nuclear types that exist in the input images. In each domain, the prototype features $F_{pt}^{c}$ for the class $c$ pass through the corresponding adversarial discriminators $D_{c}$ for class-aware adaption at the feature level.

To avoid manually finetuning the trade-off weights for the adversarial loss in N categories of the NC branch, we let the overall framework automatically learn these weight parameters during training. The learnable weighted discriminator loss is formulated as follows:

\begin{equation}
\mathcal{L}_{NC}^{ca} = \omega_c^L\sum_{c=1}^{}\mathcal{L}^{adv}_{c}
\end{equation}
where the $\mathcal{L}^{adv}_{c}$ denotes the adversarial training loss of $D_{c}$ for class $c$, and $\omega_c^L$ is its corresponding learnable loss weight.

The overall domain discriminator loss function of our model is deﬁned as:

\begin{equation}
\mathcal{L}_{dis} = \mathcal{L}_{NC}^{ca}  +\mathcal{L}_{NP}^{adv}  +\mathcal{L}_{HV}^{adv} 
\end{equation}
where $\mathcal{L}_{NP}^{adv}$ and $\mathcal{L}_{HV}^{adv}$ are the feature adaptation loss functions in the NP and HV branches, respectively. Particularly, we utilize adversarial domain discriminators on the cross-domain output features of the NP and HV branches for adaptation. Details of the supervised Hover-Net loss and the adversarial loss are shown in the Appendices. With the above loss terms, the overall loss function of the first stage approach can be written as:

\begin{equation}
\mathcal{L}_{s1} = \mathcal{L}_{F} + \mathcal{L}_{dis}
\end{equation}

\subsection{Nuclei-level Prototype Pseudo-labelling}

Although the class-aware feature alignment modules can narrow the domain gaps, the lack of supervised optimization on the target images still limits the model's segmentation and classification performance. Therefore, in the second stage, we use the output of the first stage model as pseudo labels for self-supervised learning to further improve the model performance on the target images. The detailed training process can be referred to Fig.~\ref{fig1}b.

Different from the traditional self-training process with the pseudo labels, we only train the target domain of the HV branch during the second stage pseudo-labelling process. In the extensive experiments, we noticed that the performance of the classification predictions on the target images from the first stage model is limited. In addition, the binary segmentation predictions lack object-wise information. Moreover, the model trained with all pseudo labels may not perform well due to the low quality of some pseudo-labelling classes. To avoid the disturbance from the less accurate and representative pseudo labels, we no longer consider classification and binary segmentation branches in the second stage but particularly focus on the predictions from the HV branch, where the feature maps describe the distance from each pixel to the nuclei's centre point. Therefore, the features for each nuclear object can be regarded as a prototype at the object level, which contains morphological information such as the shape and size of the specific nuclear. By self-supervised learning with the pseudo labels on the predictions of the HV branch, the bias between nuclei objects can be further reduced. The overall loss function for the second stage is as follows:

\begin{equation}
\mathcal{L}_{p} = \frac{1}{N^p}\sum_{i=1}^{N^p}|x_i^p - \hat{y}_i^p|^2
\end{equation}
where $x_i^p$ is the predicted features from the HV branch for each nuclear object $p$ in the second stage and $\hat{y}_i^p$ is the object features generated by the pseudo labels.

\section{Experiments}
\subsection{Datasets and Evaluation Metrics}
We conduct experiments on two datasets from Lizard \cite{18}, a large-scale colon tissue histopathology database at the 20x objective magnification for nuclei instance segmentation and classification under six types: epithelial, connective tissue, lymphocytes, plasma, neutrophils, and eosinophils.

In this work, DigestPath (Dpath) and CRAG are employed, where the images in Dpath are extracted from histological samples from four different hospitals in China, and the CRAG dataset contains images extracted from whole-slide images (WSIs) from University Hospitals Coventry and Warwickshire (UHCW). For both two datasets, we select 2/3 of the whole images for training, and the remaining 1/3 for testing and validation. Specifically, we use Dpath as the source domain, with 46 images for training and the rest 22 for validation. CRAG dataset is used as the target domain, with 42 images for training and the remaining 21 for testing. The training images are randomly cropped to 256×256 patches, and the data augmentation methods are applied, including flip, rotate, Gaussian blur and median blur.

For evaluation, we choose the same metrics as Hover-Net \cite{6}. Dice, Aggregated Jaccard Index (AJI), Detection Quality (DQ), Segmentation Quality (SQ), and Panoptic Quality (PQ) are for nuclei instance segmentation. In addition, the F1-scores at the detection and classification levels are employed to evaluate the nuclei detection and classification performance.

\subsection{Implementation Details}

We utilize the Hover-Net framework with ResNet50 \cite{19} pre-trained weights on ImageNet as our base architecture. In the first stage of our adaptation process, the model is trained in two steps following the Hover-Net~\cite{6}. In the first step, only the decoders are trained 50 epochs. In the second step, all layers are trained for another 50 epochs. We use Adam optimization in both steps, with an initial learning rate of 1e-4, which is then reduced to 1e-5 after 25 epochs. In the self-training pseudo-labelling part, the Adam optimizer with a learning rate of 1e-4 was used to train 50 epochs with a batch size of 20. Experiments were conducted on one NVIDIA GeForce 3090 GPU and implemented using PyTorch.

\subsection{Comparison Experiments}

We conduct a series of comparative experiments to compare the performance. The details are as follows: (1) Source Only \cite{6}: original Hover-Net without adaptation. (2) PDAM \cite{8}: a UDA nuclei instance segmentation method with pixel-level and feature-level adaptation. Since it can only be used for binary classiﬁcation for objects, we only compare the results of nuclei instance segmentation with this method. (3) Yang et al. \cite{9}: a UDA framework is proposed based on global-level feature alignment for nuclei classification and nuclear instance segmentation. In addition, a weakly-supervised DA method is also proposed using weak labels for the target images such as nuclei centroid. We only compare with the UDA method in this work for a fair comparison. (4) Fully-supervised: fully supervised training on the labelled target images, as the upper bound.

\subsubsection{Evaluation on Nucleus Instance Segmentation.}

\begin{table}
\centering
\caption{Experimental results on UDA nuclei instance segmentation.}\label{tab1}

\begin{tabular}{|l|l|l|l|l|l|}
 \hline
  &    \multicolumn{5}{l|}{$Dpath \to CRAG$}   \\
   \hline

{\bfseries Methods} &  {\bfseries Dice} & {\bfseries AJI} & {\bfseries DQ}  & {\bfseries SQ}  & {\bfseries PQ}    \\
  \hline
{\bfseries Source Only \cite{6}} & 0.378
& 0.201 &  0.336 &  0.768 &  0.259  \\
{\bfseries PDAM \cite{8}} & 0.596  & 0.323 & 0.467  & 0.676 & 0.316    \\
{\bfseries Yang {\itshape et al.} \cite{9} } & 0.766  & 0.494 &0.648   &0.765  & 0.496   \\
  \hline
{\bfseries Baseline} &  0.750  & 0.455 & 0.604 &  0.759 &  0.458  \\
{\bfseries Baseline+CA } &  0.772  & 0.502 & 0.661 & {\bfseries 0.773}  &  0.510 \\
{\bfseries Baseline+CA+PL } & 0.781    & 0.517 &  0.675 &  0.772 & 0.522 \\
{\bfseries Proposed} &  {\bfseries 0.785}  & {\bfseries 0.519} & {\bfseries 0.681} &   0.769 &  {\bfseries 0.524}  \\
{\bfseries Full-supervised} &   0.778  & 0.526 & 0.683 &  0.783 &  0.535  \\

\hline
\end{tabular}
\end{table}

A comparison of the segmentation performance between our model and state-of-the-art methods is reported in Table~\ref{tab1}. From the table, it can be observed that our instance segmentation effect is better than the two existing models. Compared with the source-only method, PDAM \cite{8} has a performance improvement by aligning features at the panoptic level. The UDA method of Yang et al. \cite{9} achieves good segmentation performance. Our method achieves the highest scores among all methods, with Dice and AJI being 2.3\% and 1.6\% higher than the previous methods, respectively. In comparison with the full-supervised model, the performance of our proposed UDA architecture is close to it. In particular, our Dice is higher than the results of the full-supervised model.

\subsubsection{Evaluation on Nucleus Classification.}

\begin{table}
\centering
\caption{Experimental results on UDA nuclei classification. $F^1_c$, $F^2_c$, $F^3_c$, $F^4_c$, $F^5_c$ and $F^6_c$ denote the F1 classiﬁcation score for the Eosinophil, Epithelial, Lymphocyte, Plasma, Neutrophil and Connective tissue, respectively. $F_{avg}$ denotes the average of all the F1-score for the classification under each category.}\label{tab2}
\begin{tabular}{|l|l|l|l|l|l|l|l|l|l|l|}
 \hline
  &    \multicolumn{8}{l|}{$Dpath \to CRAG$}   \\
   \hline

{\bfseries Methods} &  {\bfseries Det} & {\bfseries $F^1_c$} & {\bfseries $F^2_c$}  & {\bfseries $F^3_c$}  & {\bfseries $F^4_c$}  & {\bfseries $F^5_c$}  & {\bfseries $F^6_c$} & {\bfseries $F_{avg}$} \\
  \hline
{\bfseries Source Only \cite{6}} & 0.490 & 0.022 &  0.389 &  0.324 &  0.195 &  0.038 &  0.161 &  0.188\\
{\bfseries Yang {\itshape et al.} \cite{9}} &  0.736 & 0.037 &  0.670 &  0.330 &  0.371 &  0.017 &  0.428& 0.309 \\
\hline
{\bfseries Baseline} &  0.702  & 0.044 & 0.687 &  0.292 &  0.381 & 0.155 & 0.475 & 0.339 \\
{\bfseries Baseline+CA } &  0.731  & {\bfseries 0.128} & 0.697 &  0.351 &  0.400 & 0.084 & 0.498 & 0.360 \\
{\bfseries Baseline+CA+PL } &  0.772   & 0.110 &  {\bfseries0.725} &  0.327 & 0.383 & 0.352 & {\bfseries0.558} &  0.409\\
{\bfseries Proposed} &  {\bfseries 0.775} & 0.111 & 0.714 &  {\bfseries 0.389} &  {\bfseries 0.402} &   {\bfseries0.377} &   0.535 &  {\bfseries0.421}\\
{\bfseries Full-supervised} &  0.748   & 0.167 &  0.724&   0.388 &  0.419 &  0.428 &  0.545 & 0.445 \\

\hline
\end{tabular}
\end{table}

Table~\ref{tab2} reports the performance comparison of our method with other works on nuclei classification. Although Yang et al. \cite{9} can achieve cross-domain nuclei classification and segmentation based on Hover-Net, their feature alignment modules are class-agnostic, which incurs the misalignment issues for the cross-domain features from different categories and further limits the performance. On the other hand, our proposed class-aware UDA framework has outperformed \cite{9} by a large margin. Our model achieves a 4\% improvement in the F1 detection score in the detection metric. In addition, our method improves the classification performance for all categories. Moreover, our method achieves significant improvements in classes with sparse samples like Neutrophil and Eosinophil, around 10\% and 34\%, respectively. 

\subsection{Ablation Studies} 

To test the validity of each component of our proposed model, we conducted ablation experiments. Firstly, based on our architecture, we kept only the feature-level domain discriminator on the three branches as our baseline method. Secondly, we kept only the class-aware structure, removing the pseudo labels based on the nuclei-level prototype (Baseline+CA). In addition, we also compared our nuclei-prototype pseudo-labelling process with the traditional one, which directly trains the model on the target images with all predictions in the first stage as the pseudo labels (Baseline+CA+PL).

Table~\ref{tab1} and Table~\ref{tab2} show the instance segmentation and classification performance of all the ablation methods. From the tables, we can observe that the class-aware structure substantially improves the classification performance under categories with sparse samples (e.g. Neutrophil and Eosinophil), with a higher than 10\% improvement in the classification F1-score. This phenomenon illustrates the effectiveness of class-aware adaptation in transferring the knowledge between the multi-class datasets. We note that the class-aware structure also has an approximate 2\% improvement in nuclei segmentation.

\begin{figure}
\center{\includegraphics[scale=0.15]{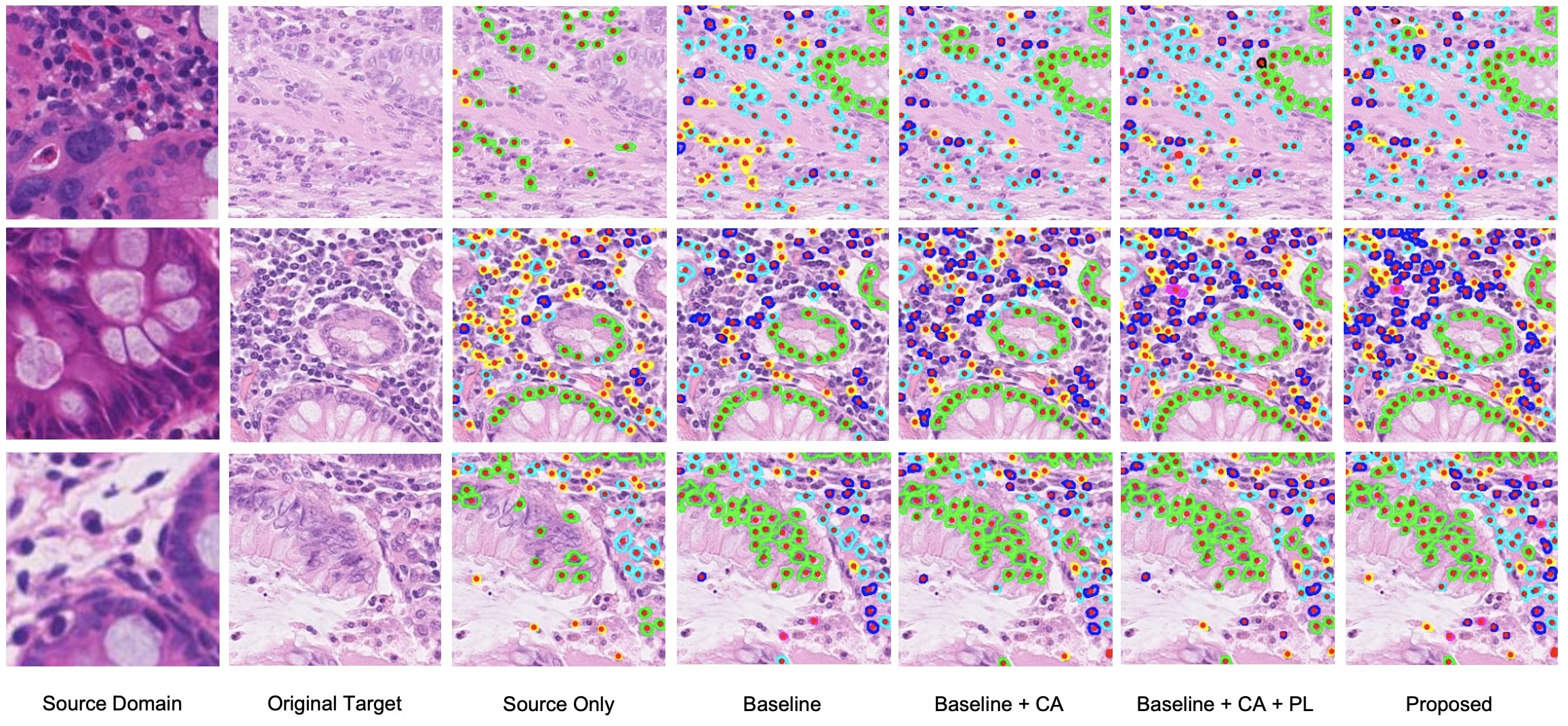}}
\caption{Visualization predictions for the ablation experiments. Red: Eosinophil, Green: Epithelial, Yellow: Lymphocyte, Blue: Plasma, Magenta: Neutrophil, Cyan: Connective tissue. (Best to viewed in color and zoomed-in)} \label{fig2} \label{framework}
\end{figure}

Self-supervised training also improved both instance segmentation and classification performance. In addition, we note that prototype loss has a higher than 4\% improvement on the F1-score for nuclei detection and achieves a better nuclei segmentation performance. In addition, our proposed nuclei prototype pseudo-labelling process also outperforms the typical pseudo-labelling. Due to the inferior classification performance of the first stage model, training models with all the pseudo labels might bring the noise to the network optimization, and limit the overall performance. Visualization examples of the ablation studies are shown in Fig.~\ref{fig2}.

\section{Conclusion} 

In this paper, we proposed a category-aware prototype pseudo-labelling architecture for unsupervised domain adaptive nuclear instance segmentation and classification. In our two-stage framework, category-aware feature alignment with learnable trade-off loss weights is proposed to tackle the class-imbalance issue and avoid misalignment during the cross-domain study. In addition, we proposed a nuclei-level prototype loss to correct the deviation in the second stage pseudo-labelling training, which further improves the segmentation and classification performance on the target images by introducing auxiliary self-supervision. Comprehensive results on various cross-domain nuclei instance segmentation and classification tasks demonstrate the prominent performance of our approach. Given the appealing performance of our method on the UDA nuclei instance segmentation tasks, we suggest that future directions can focus on the cross-domain multi-class object recognition tasks for other medical and general computer vision scenarios.


\clearpage
	\newpage
	\section*{{\Large Appendices}}
	\label{sec_appendix}
	\appendix
	\input{appendix}

\end{document}

%% file: appendix.tex
\section{Summary of the annotations in our experiment}
\begin{table}
\centering
\caption{Details of the number of nuclei under each class in the datasets we used, DigestPath, CRAG and GlaS.}\label{tab1}
\begin{tabular}{|l|l|l|l|l|}
\hline
{\bfseries Class} &  {\bfseries DigestPath} & {\bfseries CRAG} & {\bfseries GlaS} \\
\hline
{\bfseries Epithelial} &  70,789 & 99,124 & 31,986\\
{\bfseries Lymphocyte} &  49,932 & 27,634 & 9,763\\
{\bfseries Plasma} &  11,352 & 9,363 & 2,349\\
{\bfseries Neutrophil} &  2,262 & 1,673 & 90 \\
{\bfseries Eosinophil} &  1,349 & 1,255 & 286\\
{\bfseries Connective} &  32,826 & 49,994 & 10,890\\
{\bfseries Total} &  168,510 & 189,043 & 55,364\\
\hline
\end{tabular}
\end{table}

% \section{Experimental results on Dpath to Glas dataset}

\section{Loss function }
\subsection{The supervised Hover-net loss}
The supervised Hover-Net loss function of our model is deﬁned as $\mathcal{L}_F$:
\begin{equation}
\mathcal{L}_{F}= \mathcal{L}_{n p}+\mathcal{L}_{hover}+\mathcal{L}_{nc}
\end{equation}

where $\mathcal{L}_{np}$, $\mathcal{L}_{hover}$ and $\mathcal{L}_{nc}$ represent the loss with respect to the output at the NP, HV and NC branch, respectively.

For the NP and the NC branches, the loss is calculated by adding cross-entropy loss and dice loss:

\begin{equation}
\mathcal{L}_{n p}=\mathcal{L}_{n p}^{C E}+\mathcal{L}_{n p}^{dice }
\end{equation}

\begin{equation}
\mathcal{L}_{n c}=\mathcal{L}_{n c}^{C E}+\mathcal{L}_{n c}^{dice }
\end{equation}

The cross entropy and dice losses are defined as:

\begin{equation}
\mathcal{L}^{C E}=-\frac{1}{N} \sum_{i=1}\left[p_{i} \log \left(q_{i}\right)\right]
\end{equation}

\begin{equation}
\mathcal{L}^{dice }=1-\frac{2\left(q_{i} * p_{i}\right)}{q_{i}+p_{i}}
\end{equation}

For the HV branch, we denoted $\mathcal{L}^{mqe}$ as the mean squared error loss, and defined the $\mathcal{L}^{hv}$ loss as the mean squared error between the horizontal and vertical gradients and the corresponding gradients of the ground truth:

\begin{equation}
\mathcal{L}_{hover }=\mathcal{L}^{m q e}+\mathcal{L}^{h v}
\end{equation}

\begin{equation}
\mathcal{L}^{m q e}=\frac{1}{N} \sum_{i=1}\left(p_{i}-q_{i}\right)^{2}
\end{equation}

\begin{equation}
\mathcal{L}^{h v}=\frac{1}{M} \sum_{i=1}\left(p_{i, h o r}-q_{i}\right)^{2}-\frac{1}{M} \sum_{i=1}\left(p_{i, v e r}-q_{i}\right)^{2}
\end{equation}

\subsection{The domain discriminator loss}

The overall domain discriminator loss function of our model is deﬁned as $\mathcal{L}_{dis}$: 

\begin{equation}
\mathcal{L}_{dis} = \mathcal{L}_{NC}^{ca}  +\mathcal{L}_{NP}^{adv}  +\mathcal{L}_{HV}^{adv} 
\end{equation}

where $\mathcal{L}_{NC}^{ca}$, $\mathcal{L}_{NP}^{adv}$ and $\mathcal{L}_{HV}^{adv}$ are the features adaptation loss functions in the NC, NP and HV branches, respectively.

The learnable weighted discriminator loss is formulated as follows: 

\begin{equation}
\mathcal{L}_{NC}^{ca} = \omega_c^L\sum_{c=1}^{}\mathcal{L}^{adv}_{c}
\end{equation}

where the $\mathcal{L}^{adv}_{c}$ denotes the adversarial training loss of $D_{c}$ for class $c$, and $\omega_c^L$ is its corresponding learnable loss weight.

Concretely, we deﬁne the adversarial training loss as:

\begin{equation}
\mathcal{L}^{a d v}=-\frac{1}{N} \sum_{i=1}\left[y_{i} \log \left(p_{i}\right)+\left(1-y_{i}\right) \log \left(1-p_{i}\right)\right]
\end{equation}

With the above loss terms, the overall loss function of the first stage approach can be written as:

\begin{equation}
\mathcal{L}_{s1} = \mathcal{L}_{F} + \mathcal{L}_{dis}
\end{equation}

\subsection{The prototype pseudo-labelling loss}

The prototype pseudo-labelling loss function for the second stage is as follows:

\begin{equation}
\mathcal{L}_{p} = \frac{1}{N^p}\sum_{i=1}^{N^p}|x_i^p - \hat{y}_i^p|^2
\end{equation}
where $x_i^p$ is the predicted features from the HV branch for each nuclear object $p$ in the second stage and $\hat{y}_i^p$ is the object features generated by the pseudo labels.

\newpage
\section{Supplement on the experimental results}

\begin{table}
\centering
\caption{Experimental results on UDA nuclei instance segmentation by transferring from $Dpath$ to the $Glas$ dataset. }\label{tab1}

\begin{tabular}{|l|l|l|l|l|l|}
 \hline
  &    \multicolumn{5}{l|}{$Dpath \to GlaS$}   \\
   \hline

{\bfseries Methods} &  {\bfseries Dice} & {\bfseries AJI} & {\bfseries DQ}  & {\bfseries SQ}  & {\bfseries PQ}    \\
  \hline
{\bfseries Source Only } & 0.494   & 0.264 &  0.359 &  0.732 &  0.262  \\
{\bfseries PDAM } & 0.571  &  {\bfseries 0.298} & 0.400  & 0.667 & 0.267   \\
{\bfseries Yang {\itshape et al.} } & 0.639    & 0.294 & 0.377  &0.735 &  0.275   \\
  \hline
{\bfseries Baseline} &  0.645    & 0.288 & 0.374 &  {\bfseries 0.739} &  0.275  \\
{\bfseries Proposed} &  {\bfseries 0.651}   & 0.296 & {\bfseries 0.385} &   0.734 & {\bfseries 0.281}  \\
{\bfseries Full-supervised} &   0.721  &  0.423 & 0.539 &   0.765 &  0.411  \\

\hline
\end{tabular}
\end{table}

\begin{table}
\centering
\caption{Experimental results on UDA nuclei classification under the UDA scenario: $Dpath \to GlaS$}\label{tab2}
\begin{tabular}{|l|l|l|l|l|l|l|l|l|l|l|}
 \hline
  &    \multicolumn{8}{l|}{$Dpath \to GlaS$}   \\
   \hline

{\bfseries Methods} &  {\bfseries Det} & {\bfseries $F^1_c$} & {\bfseries $F^2_c$}  & {\bfseries $F^3_c$}  & {\bfseries $F^4_c$}  & {\bfseries $F^5_c$}  & {\bfseries $F^6_c$} & {\bfseries $F_{avg}$} \\
  \hline
{\bfseries Source Only } & 0.512  & {\bfseries 0.022} &  0.467 &  {\bfseries 0.206} &  0.068 &  {\bfseries 0.215} &  0.186 &  0.194\\
{\bfseries Yang {\itshape et al.}} &  0.565  & 0.000 &  0.621 &   0.141 &   0.152 & 0.018 &  0.240 & 0.195  \\
\hline
{\bfseries Baseline} &  0.566     & 0.000 & 0.617 &  0.145 &  0.175 & 0.063 & 0.249 & 0.208 \\
{\bfseries Proposed} &  {\bfseries 0.572}   & 0.000 &  {\bfseries 0.647}  &  0.131 &   {\bfseries 0.192} &    0.168 &  {\bfseries 0.323} & {\bfseries 0.244}\\
{\bfseries Full-supervised} &  0.675   &  0.000 &  0.775&   0.207 &   0.191 &  0.000 &  0.440 & 0.269 \\

\hline
\end{tabular}
\end{table}

\begin{table}
\centering
\caption{Computational complexity analysis, which indicates our proposed modules can introduce performance gain but bring negligible auxiliary cost.}\label{tab1}

\begin{tabular}{|l|l|l|}
 \hline
  &    \multicolumn{2}{l|}{$Dpath \to GlaS$}   \\
   \hline

{\bfseries Methods} &  {\bfseries Time cost} & {\bfseries Number of Parameters}  \\
  \hline
{\bfseries Source Only } &  35 s/iter & 3.7M  \\
{\bfseries Baseline} &   45 s/iter & 3.7M   \\
{\bfseries Proposed} &   45 s/iter & 3.7M  \\

\hline
\end{tabular}
\end{table}

\newpage

% \section{Visualization predictions for the experiment}
\section{Visualization predictions for Dpath to GlaS experiment}
\begin{figure}
\centering
\center{\includegraphics[scale=0.18]{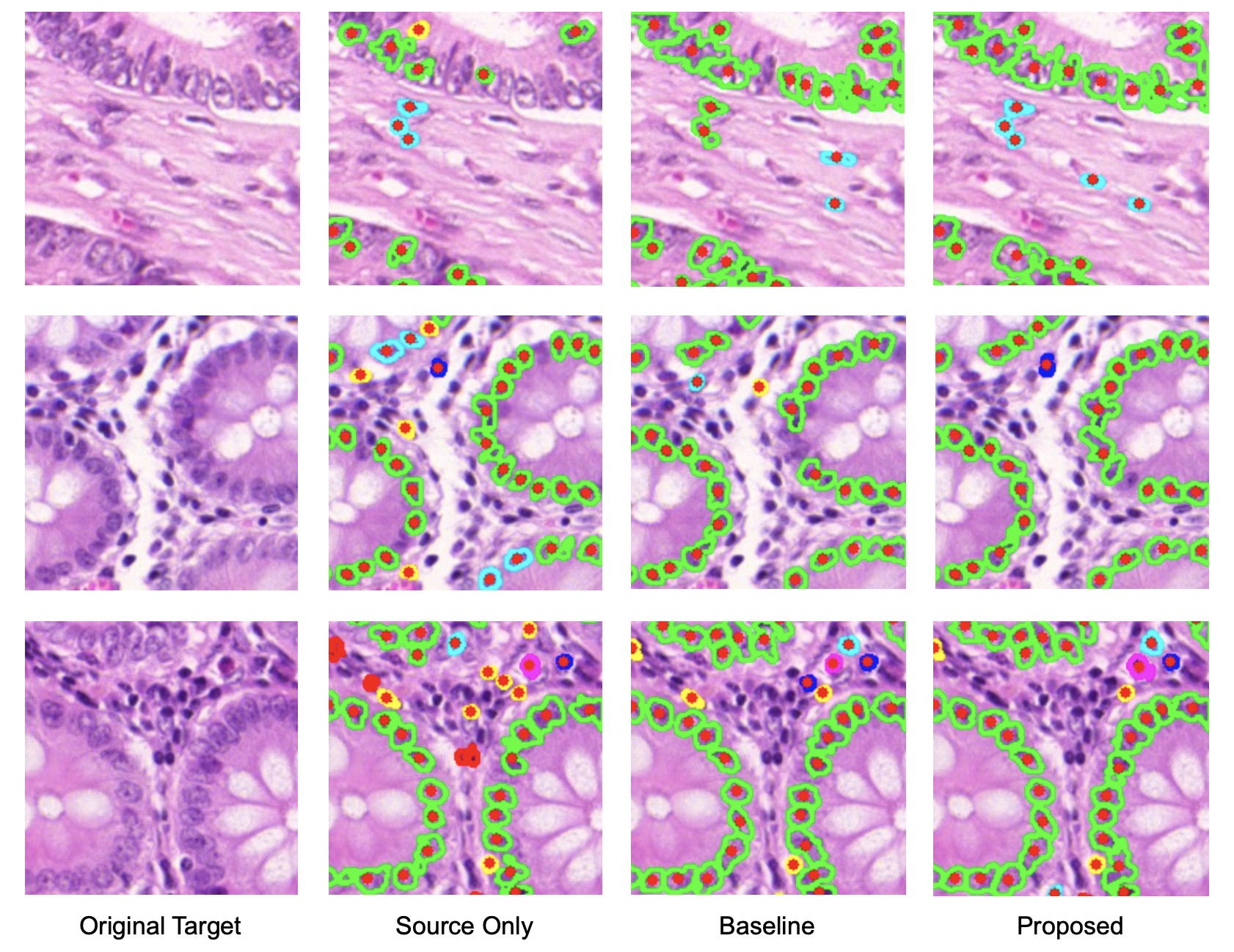}}
\caption{Visualization predictions for the UDA experiment under the $Dpath \to GlaS$ setting.} \label{fig2} \label{framework}
\end{figure}

\section{Limitations}
There are several limitations in our proposed method:

1) The pseudo labels for the nuclei prototype still contain noises, which limits the self-supervised learning performance.

2) The current method might incur performance drop under some specific cross-domain nuclei instance segmentation and classification tasks, such as from Glas to Dpath.